\documentclass{esannV2}
\usepackage[dvips]{graphicx}
\usepackage[latin1]{inputenc}
\usepackage{amssymb,amsmath,array}
\usepackage{hyperref}
\usepackage{longtable}
\usepackage{makecell}
\usepackage{pdfpages}

\begin{document}
\title{The Top 10 Topics in Machine Learning Revisited: A Quantitative Meta-Study}

\author{Patrick Glauner$^1$, Manxing Du$^1$, Victor Paraschiv$^2$, Andrey Boytsov$^1$,  \\ Isabel L\'{o}pez Andrade$^3$, Jorge Augusto Meira$^1$, Petko Valtchev$^{14}$ and Radu State$^1$
%
%
\vspace{.3cm}\\
%
1- Interdisciplinary Centre for Security, Reliability and Trust, University of Luxembourg \\
4 rue Alphonse Weicker, 2721 Luxembourg, Luxembourg
%
\vspace{.1cm}\\
2- Numbers of others \\
London, United Kingdom
\vspace{.1cm}\\
3- American Express \\
Sussex House, Civic Way, Burgess Hill RH15 9AQ, United Kingdom
\vspace{.1cm}\\
4- Department of Computer Science, University of Quebec in Montreal \\
201, av. President Kennedy, Montreal H2X 3Y7, Canada
}
\setlength{\tabcolsep}{0pt}

\maketitle

\begin{abstract}
Which topics of machine learning are most commonly addressed in research? This question was initially answered in 2007 by doing a qualitative survey among distinguished researchers. In our study, we revisit this question from a quantitative perspective. Concretely, we collect 54K abstracts of papers published between 2007 and 2016 in leading machine learning journals and conferences. We then use machine learning in order to determine the top 10 topics in machine learning. We not only include models, but provide a holistic view across optimization, data, features, etc. This quantitative approach allows reducing the bias of surveys. It reveals new and up-to-date insights into what the 10 most prolific topics in machine learning research are. This allows researchers to identify popular topics as well as new and rising topics for their research.
\end{abstract}

\section{Introduction}
In 2007, a paper named ``Top 10 algorithms in data mining" identified and presented the top 10 most influential data mining algorithms within the research community \cite{Wu2008}. The selection criteria were created by consolidating direct nominations from award winning researchers, the research community opinions and the number of citations in Google Scholar. The top 10 algorithms in that prior work are: C4.5, k-means, support vector machine, Apriori, EM, PageRank, AdaBoost, kNN, naive Bayes and CART.

In the decade that passed since then, machine learning has expanded, responding to incremental development of computational capabilities and substantial increase of problems in the commercial applications. This study reflects on the top 10 most popular fields of active research in machine learning, as they emerged from the quantitative analysis of leading journals and conferences. This work sees some topics in the broader sense including not only models but also concepts like data sets, features, optimization techniques and evaluation metrics. This wider view on the entire machine learning field is largely ignored in the literature by keeping a strong focus entirely on models \cite{domingos2012few}.

Our core contribution in this study is that we provide a clear view of the active research in machine learning by relying solely on a quantitative methodology without interviewing experts. This attempt aims at reducing bias and looking where the research community puts its focus on. The results of this study allow researchers to put their research into the global context of machine learning. This provides researchers with the opportunity to both conduct research in popular topics and identify topics that have not received sufficient attention in recent research.
The rest of this paper is organized as follows. Section~\ref{section:top10} describes the data sources and quantitative methodology. Section~\ref{section:res} presents and discusses the top 10 topics identified. Section~\ref{section:conclusions} summarizes this work.

\section{Methodology}
\label{section:top10}
In this section, we discuss how we determine quantitatively the top 10 topics in machine learning from articles of leading journals and conferences published between January 2007 and June 2016. We selected referenced journals that cover extensively the field of machine learning, neural networks, pattern recognition and data mining both from the theoretical perspective and also with applications on image, video and text processing, inference on networks and graphs, knowledge basis and applications on real data sets. 

\subsection{Data collection}
In the data collection, we focus on the abstracts of publications, as they provide the main results and conclusions of a paper. In contrast, the full text includes details on the research, which also comes with more noise that is not relevant to an overall summary of published work.
We have chosen 31 leading journals related to machine learning as summarized in Table~\ref{table:journals}, ranked by their impact factor.
For each journal, we have collected as many abstracts as possible of articles published in the timeframe of interest. In total, we have collected 39,067 abstracts of those 31 journals, which also include special issues.

\renewcommand{\arraystretch}{0.5}
\begin{longtable}{|l|c|c|}
\hline
Name & Imp. Fct. & \#Abstr.\\
\hline
IEEE T. on Sys., Man, and Cybernetics, P. B. (Cyb.) & 6.22 & 1,045 \\
IEEE T. on Pattern Analysis and Machine Intell. & 5.781 & 2,552 \\
IEEE T. on Neural Networks and Learning Systems & 4.291 & 1,518 \\
IEEE T. on Evolutionary Computation & 3.654 & 940 \\
IEEE T. on Medical Imaging & 3.39 & 2,470 \\
Artificial Intelligence & 3.371 & 668 \\
ACM Computing Surveys & 3.37 & 395 \\
Pattern Recognition & 3.096 & 3,016 \\
Knowledge-Based Systems & 2.947 & 1,905 \\
Neural Networks & 2.708 & 1,330 \\
IEEE T. on Neural Networks & 2.633 & 758 \\
IEEE Computational Intelligence Magazine & 2.571 & 574 \\
IEEE T. on Audio, Speech and Language Processing & 2.475 & 1,829 \\
Journal of Machine Learning Research & 2.473 & 986 \\
IEEE Intelligent Systems & 2.34 & 1,049 \\
Neurocomputing & 2.083 & 6,165 \\
IEEE T. on Knowledge and Data Engineering & 2.067 & 2,121 \\
Springer Machine Learning & 1.889 & 571 \\
Computer Speech and Language & 1.753 & 452 \\
Pattern Recognition Letters & 1.551 & 2,380 \\
Computational Statistics \& Data Analysis & 1.4 & 3,063 \\
Journal of the ACM & 1.39 & 353 \\
Information Processing \& Management & 1.265 & 730 \\
ACM T. on Intelligent Systems and Technology & 1.25 & 396 \\
Data \& Knowledge Engineering & 1.115 & 660 \\
ACM T. on Information Systems & 1.02 & 229 \\
ACM T. on Knowledge Discovery from Data & 0.93 & 245 \\
ACM T. on Autonomous and Adaptive Systems & 0.92 & 231 \\
ACM T. on Interactive Intelligent Systems & 0.8 & 117 \\
ACM T. on Applied Perception & 0.65 & 234 \\
ACM T. on Economics and Computation & 0.54 & 85 \\
\hline
Total (N=31) & - & 39,067 \\
\hline
\caption{Source journals.}
\label{table:journals}
\end{longtable}

Furthermore, we have chosen 7 major international conferences related to machine learning as summarized in Table~\ref{table:conferences}, ranked by their average citation count. We have collected as many proceedings as possible of those conferences.
\begin{table}[h!]
\begin{minipage} {\textwidth}
\centering
\renewcommand{\arraystretch}{0.5}
\begin{tabular}{|l|c|c|c|}
\hline
Name & \#Avg. Cit. & \#Abstr. & Years\\
\hline
Inter. Conference on Computer Vision & 11.9754 & 2,092 & \makecell{2007, 2009, \\ 2011, 2013,\\ 2015}\\
Inter. Conference on Machine Learning & 9.1862 & 1,185 & 2013-2016 \\
Advs. in Neural Information Processing Syst. & 8.5437 & 2,416 & 2007-2015 \\
Conf. on Knowledge Discovery and Data M. & 7.7269 & 1,035 & 2007-2015 \\
Conf. on Computer Vision and Pattern Recog. & 6.6133 & 4,471 & 2007-2015 \\
Conference on Learning Theory & 4.2905 & 347 & 2011-2016 \\
International Conference on Data Mining & 2.137 & 1,406 & 2007-2015 \\
J. of Machine Learning Research Conf. Proc. & 2.473\footnote{Computing the average citation count of this mixture of various conferences and workshops has proven to not be feasible. Instead, we use the impact factor of the Journal of Machine Learning Research as the average citation count. We expect the impact of the approximation error to be low since it only concerns 1,507 of the total 53,526 abstracts used in this research.}
& 1,507 & 2007-2016 \\
\hline
Total (N=8) & - & 14,459  & - \\
\hline
\end{tabular}
\end{minipage} 
\caption{Source conferences.}
\label{table:conferences}
\end{table}
In addition, we consider the Journal of Machine Learning Research Workshop and Conference Proceedings series, which includes further conferences, such as International Conference on Artificial Intelligence and Statistics and Asian Conference on Machine Learning among others.
We have collected 14,459 abstracts from the proceedings of those conferences in the time frame of interest. Combining the journals and conference proceedings, we have collected 53,526 abstracts in total.

\subsection{Key phrase extraction}
We focus on extracting the most relevant key phrases of each abstract, which we call \textit{topics} in the remainder of this study.
First, we apply Porter stemming to an abstract \cite{Porter:1997:ASS:275537.275705}. In stemming, only the stem of a word is retained. For example, ``paper" and ``papers" have the same stem, which is ``paper".
For the extraction of key phrases from each abstract, we compare two different methods:

\begin{enumerate}
\itemsep -.5em 
\item We remove the stop words\footnote{Stop words are the words most frequently used in a language that usually provide very little information. For example, ``and" or "of" are typical stop words in the English language.} from each abstract and then use all bigrams and trigrams as key phrases.
\item The Rapid Automatic Keyword Extraction Algorithm (RAKE) is an unsupervised, domain-independent and language-independent learning algorithm that generates key phrases from text \cite{rakerose}. First, RAKE splits each abstract into parts that are separated by signs - such as commas and full stops - and stop words. These parts are then split into n-gram key phrases. In our work, we use $1 \le n \le 4$. Next, a word co-occurrence graph is learned from the generated n-grams. Last, each key phrase is ranked by the sum of the ratio of degree to frequency per word.
\end{enumerate}

When merging the key phrases of different journals or conferences, we weight each key phrase by the impact factor or average citation count, respectively. The list of key phrases is then sorted in descending order by their total weighted count.
We then manually clean the top 500 key phrases by removing key phrases unrelated to machine learning, such as ``propos[ed]\footnote{Stemmed words are completed to their original form for clarity in this paper.} method" or ``experiment[al] result[s] show", but also other irrelevant computer science terms, such as ``comput[er] vision". Last, starting with the most popular key phrase, we iteratively skip related key phrases. We continue this merger until we find 10 distinct key phrases of different topics, which are the top 10 topics in machine learning. For example, key phrases related to ``data set" are ``train[ing] data" and ``real data". 
Our implementation is available as open source: \url{http://github.com/pglauner/MLtop10}.



\section{Results}
\label{section:res}
\begin{figure}[h!]
    \centering
    \includegraphics[width=0.75\textwidth]{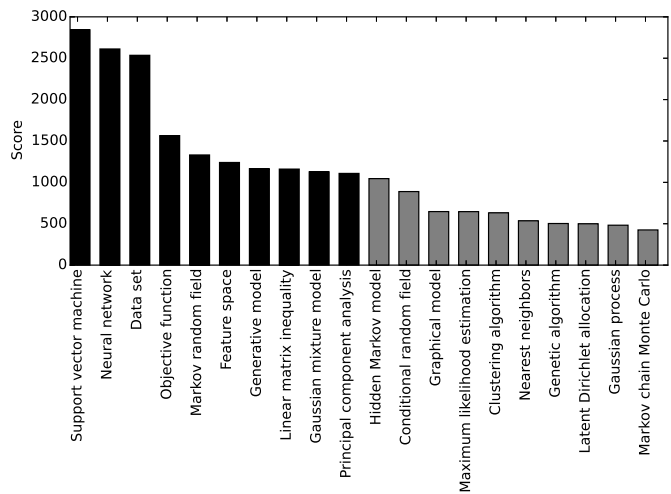}
    \caption{Top 10 topics highlighted in black, the top 11-20 topics in grey.}
    \label{fig:bar}
\end{figure}

Using method 1, which utilizes bigrams and trigrams for extraction, we only get very general topics. Concretely, the top 5 topics are ``network pretraining", ``supervised classification part", ``learn binary representation", ``unsupervised [and] supervised learning" and ``predict label [from the] input". In contrast, performing method 2, which is machine learning based key word extraction using RAKE, we get the top 10 topics depicted in Figure~\ref{fig:bar}.
We notice that after the first three topics, i.e. ``support vector machine", ``neural network" and ``data set", there is a significant drop in terms of popularity. We notice another drop after ``objective function". The next 7 topics are vey close in terms of their popularity. ``Hidden Markov model" has a popularity only slightly lower than ``principal component analysis".

\subsection{Discussion}
Comparing the two key phrase extraction methods, we see that using RAKE we obtain more robust results that reflect for example frequent keywords and unbalanced terms much better. 

Comparing our list of top 10 topics to the list of top 10 algorithms in data mining from 2007 \cite{Wu2008}, we make the following important observations: Due to their popularity in research, we have expected that support vector machines would appear in the top 10. Also, neural networks have been celebrating a comeback under the umbrella term ``deep learning" since 2006 \cite{lecun2015deep} and we therefore expected them to appear in the top 10 as well under either term. We can also confirm that Hidden Markov models have received significantly less attention in research than neural networks over the last 10 years.
We have not expected that the linear matrix inequality would appear in the top 10. However, given its importance to the theoretical foundations of the field of machine learning it is absolutely justified to appear in the top 10. Its appearance does not indicate a fallacy in our methodology.
Naive Bayes has often been described as a wide-spread baseline model in the literature. Furthermore, tree classifiers such as random forests have become popular in the literature and do not appear in the top 10 either. Both, C4.5 and CART are tree learning algorithms that were found to be among the top 10 data mining algorithms in 2007.
In terms of models, we did not expect that Markov random fields and Gaussian mixture models receive more attention than naive Bayes or tree based learning methods in current research publications.

A quantitative approach comes with a potential new bias depending on which data sources are used. Possible factors include the quality of publications and focus of each source (journal/conference). The vast majority of source abstracts are from journals and conferences that have a high impact factor or average citation count. We have made sure to include as many sources as possible that have a wide scope. In return, we have attempted to keep the number of sources with a very narrow scope to a minimum. Also, if the inclusion or omission of a specific source is questioned, this has only very little impact due to the distribution of abstracts: There are in total 39 sources (31 journals + 8 conferences). In average, a source has 1,372 abstracts or 2.56\% of all abstracts. The largest source is the Neurocomputing journal, which has 6,165 abstracts or 11.52\% of all abstracts.

\section{Conclusions}
\label{section:conclusions}
In our study, we use machine learning in order to find the top 10 topics in machine learning from about 54K abstracts of papers published between 2007 and 2016 in leading machine learning journals and conferences. Concretely, we found support vector machine, neural network, data set, objective function, Markov random field, feature space, generative model, linear matrix inequality, Gaussian mixture model and principal component analysis to be the top 10 topics.
Compared to previous work in this field from 2007, support vector machine is the only intersection of both top 10 lists. This intersection is small for the following reasons: First, we do not only consider models, but span a wider view across the entire field of machine learning also including features, data and optimization. Second, we perform a quantitative study rather than opinion-based surveying of domain experts in order to reduce the bias. Third, the models of interest have significantly changed of the last 10 years, most prominently symbolized by the comeback of neural networks under the term deep learning.
Overall, we are confident that our quantitative study provides a comprehensive view on the ground truth of current machine learning topics of interest in order to strengthen and streamline future research activities.


\begin{footnotesize}




\bibliographystyle{unsrt}
\bibliography{References}

\end{footnotesize}


\end{document}